\documentclass{article} 
\usepackage{nips15submit_e,times}
\usepackage{hyperref}
\usepackage{url}
\usepackage{graphicx}

\title{Does the ``Artificial Intelligence Clinician'' learn optimal treatment strategies for sepsis in intensive care?}

\author{Russell Jeter\thanks{These authors contributed equally to this work.} $^{~ 1}$, Christopher Josef$^{*2}$, Supreeth Shashikumar$^3$, and Shamim Nemati$^1$
\\
$^1$Department of Biomedical Informatics, Emory University, Atlanta, USA. \\
$^2$School of Medicine, Department of Surgery, Emory University, Atlanta, USA. \\
$^3$ School of Electrical and Computer Engineering, Georgia Institute of Technology, Atlanta, USA.
}

\nipsfinalcopy 

\begin{document}

\maketitle

\section{Introduction}

From 2017 to 2018 the number of scientific publications found via PubMed search using the keyword ``Machine Learning'' increased by 46\% (4,317 to 6,307). The results of studies involving machine learning, artificial intelligence (AI), and big data have captured the attention of healthcare practitioners, healthcare managers, and the public at a time when Western medicine grapples with unmitigated cost increases and public demands for accountability. The complexity involved in healthcare applications of machine learning and the size of the associated data sets has afforded many researchers an uncontested opportunity to satisfy these demands with relatively little oversight. While there is great potential for machine learning algorithms to positively impact healthcare related costs and outcomes, its deployment cannot be regarded in the same way a shipping optimization algorithm is considered at an Amazon warehouse. Those involved in the care of the infirm appreciate the incredibly high expectation for performance and more importantly the unique obligation to \emph{primum non nocere (first do no harm).}

In a recent \emph{Nature Medicine} article, ``The Artificial Intelligence Clinician learns optimal treatment strategies for sepsis in intensive care,'' \cite{komorowski_artificial_2018} Komorowski  and his coauthors propose methods to train an artificial intelligence clinician to treat sepsis patients with vasopressors and IV fluids. In this post, we will closely examine the claims laid out in this paper. In particular, we will study the individual treatment profiles suggested by their AI Clinician to gain insight into how their AI Clinician intends to treat patients on an individual level. While the work has the potential to contribute to the advancement of the field, its unqualified claims which have been repeated throughout the lay press \cite{timmer_ais_2018} promote an unrealistic capability. These claims jeopardize future prospective applications of healthcare associated machine learning algorithms and could cause great harm to patients if prospectively deployed.

\section{Terminology}
Discussion of Reinforcement Learning as applied by Komorowski et al. to the treatment of sepsis requires understanding of several key terms. Definitions of these terms are provided so that the reader can better follow our work. Readers familiar with this topic can skip this section and begin their reading at Summary.

\textbf{State}: \textit{The summary of all of a patient's clinical measurements (e.g., heart rate, leukocyte count, gender etc.) at a given point in time. In this work every 4-hour period of patient data is assigned to one of 750 unique states.}

\textbf{Action}: \textit{A combination of two intervention categories (IV fluid and vasopressor treatment). IV fluid and vasopressor treatments are each divided into 5 treatment levels for a total of 25 permutations (5x5). Though there are many types of IV fluids and vasopressors, for simplicity, the authors normalize them to construct the two intervention categories (vasopressors and IV fluids).}

\textbf{Policy}: \textit{This is the recommended action to take for a given state.  In this context, for a given patient state, the policy is the recommended amount of vasopressors or IV fluids to give.}

\textbf{Value}: \textit{The value of a policy is defined in terms of the likelihood of survival under the proposed policy. In a retrospective setting, this requires ``Counterfactual Reasoning'', i.e. what-if the patients were dosed according to the AI clinician's policy rather than the existing policy (that is, the policy that generated the observed data)?}

\textbf{Sepsis}: \textit{A dysregulated immune response to infection that results in tissue and end organ damage.}

\textbf{Hypotension}: \textit{The metabolic processes of the body require that oxygenated blood be delivered to the organs of body with sufficient pressure. Mean arterial pressures of $<$ 65 mm Hg are considered pathologic.} 

\textbf{Septic Shock}: \textit{A condition that is diagnosed when a septic patient experiences a persistently low blood pressure that is not improved by treatment with IV fluids, requires vasopressor support, and is suspected of causing elevated lactate levels greater than 2mmol/L. Vasopressors and IV fluids are the most prevalent adjuncts used to re-establish physiological MAP levels $\geq 65$.}

\begin{figure}[h]
\begin{center}
\includegraphics[width = \textwidth]{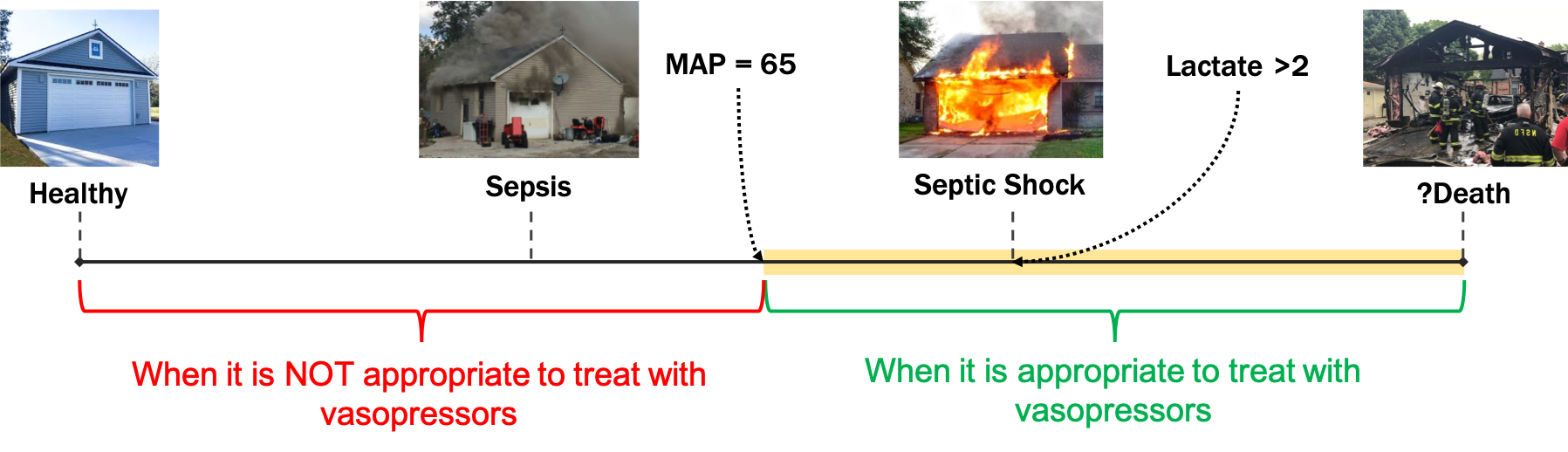}
\end{center}
\caption{There exists a small population of hypotensive patients with sepsis that requiring vasopressors, but have normal lactic acid levels. By virtue of their normal lactate they are not labeled as having septic shock.}
\label{Figure:  Sepsis Continuum}
\end{figure}

\section{Summary}
The authors approach treating septic patients using a Reinforcement Learning (RL) framework \cite{sutton_reinforcement_nodate}. This means that there is an environment (a sick patient) and an agent (a clinician) who acts to maximize some reward (patient health). Generally speaking, this is an appropriate way to automate or create ``clinician-in-the-loop'' algorithms for the treatment of a patient's experiencing an evolving clinical phenotype (state). While there are many different RL methods that can be used to approach patient treatment, Komorowski et al. chose to use a ``Policy Iteration'' method (a great introduction to this method can be found in \cite{alzantot_deep_2017}). Given sufficient information about the underlying system, this method evaluates a large state space and converges to an optimal policy: a policy that maximizes the expected value, which in this setting corresponds to minimizing patient mortality.

To utilize this method, the authors use real patient data from the public Medical Information Mart for Intensive Care III (MIMIC III) \cite{johnson_mimic-iii_2016} dataset to construct their environment and provide a clinical baseline treatment policy for comparison/evaluation. Their RL ``environment'' consists of patient features (including comorbidities) clustered into 750 states and their RL ``actions'' consist of two intervention categories (vasopressors administration and IV fluids administration) binned into 25 actions (see the Terminology section for more explanation). After the patient trajectories are compressed into these discrete states, the authors construct a 750 x 750 x 25 3-D transition matrix (or 25 different 750 x 750 transition matrices) from the observations in their dataset. Put simply, suppose your patient is in state 517, and historically, clinicians treat patients in this state with action 8. Given the current patient state 517 and clinician action 8, there is an underlying probability that the patient will transition to each of the 750 states in the environment. Unfortunately, the authors have provided no evidence of goodness of fit \cite{besag_exact_2013} for this transition matrix to the patient state data (this will be the topic of a future post).

If one accepts this representation of patient states, then using the probabilities in this transition matrix, you can predict a septic patient's entire clinical course given the treatment regimens. That is, the model provides an \emph{in silico} representation of the entire septic ICU population. The authors use this empirically-derived transition matrix to implement the policy iteration method; this enables them to construct an AI Clinician that learns how to dose patients with vasopressors and IV fluids.

\begin{figure}[h!]
\centering
\includegraphics[width = 0.6\textwidth]{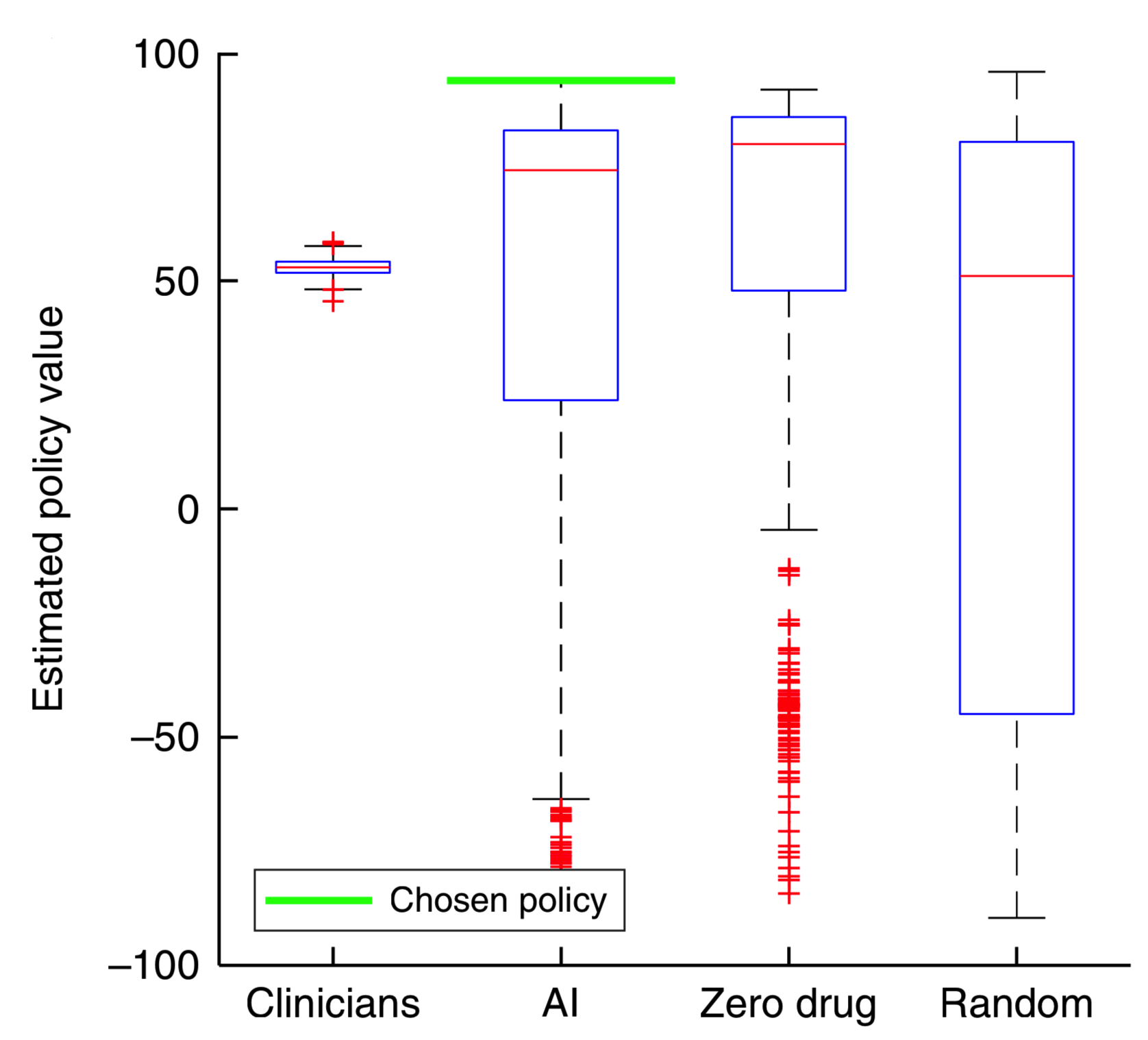}
\caption{Estimated policy value for the observed clinician policy, the learned AI policy, the zero drug policy, and a random policy for 500 realizations of the environment. Figure taken from \cite{komorowski_artificial_2018}.}
\label{Figure: Komorowski Expected Policy Value}
\end{figure}

Figure \ref{Figure: Komorowski Expected Policy Value} shows the estimated policy value for four different agents after 500 iterations whereby the environment (including states and transition probabilities) are constructed, and a new dosing policy is explored using Policy Iteration \cite{alzantot_deep_2017}. The value of the dosing policies are determined using survival or expiration as the terminal state (see equation (5) in \cite{komorowski_artificial_2018}). The chosen policy, e.g. “best”, is the one with the highest expected policy value. The results seem promising: the best AI policy yields a much higher expected value than all of the clinician policies. However, a serious concern arises too: \emph{for most of the realizations, the zero drug policy outperforms the clinician policy.} This is counterintuitive and does not pass the “smell test”.

\emph{How is it more valuable on average to NOT intervene on a population of septic patients?}

Perhaps there is an underlying problem in the proposed model or in the evaluation metric that consistently allows this zero-drug policy to outperform the clinician policy. To understand their methods in more detail, we reached out to the authors on October 31st, 2018 to obtain the code and data used in their study to train the AI Clinician (which, per their publication, is “available by request to the corresponding authors”). They responded that the code would be made available through their website, but several months later is still unavailable. Hence, we decided to reproduce their work using the MIMIC III dataset and a careful review of their methods.

\section{Results from Implementing Their Methods}
\textit{A note on reproducibility: Reproducibility is one of the cornerstones of scientific progress. It is imperative that scientists be able to reproduce someone else's findings; this allows future researchers to build on previous work as well as verify that controversial findings are valid. In computational disciplines such as biomedical informatics, making codes available via platforms such as GitHub is an important step towards reproducibility.}

To investigate why non-intervention outperformed clinician policy, we attempted to reproduce the results of Komorowski et al. We utilized data from 5,366 septic patients from the MIMIC III dataset in which every patient received both vasopressors and fluids. We chose this cohort to ensure that the patients evaluated by the AI Clinician required clinical intervention. Data was binned into one-hour intervals, whereas data in Komorowski et al.’s paper were binned into four-hour intervals. We use one-hour intervals to better approximate the dose-to-response time for fluid resuscitation and vasopressor intervention. We normalized and discretized clinician actions into the same 25 actions and similarly followed the methods described in the paper. Below is a summary of clinician actions across the cohort:

\begin{figure}[h]
\begin{center}
\includegraphics[width = 0.75\textwidth]{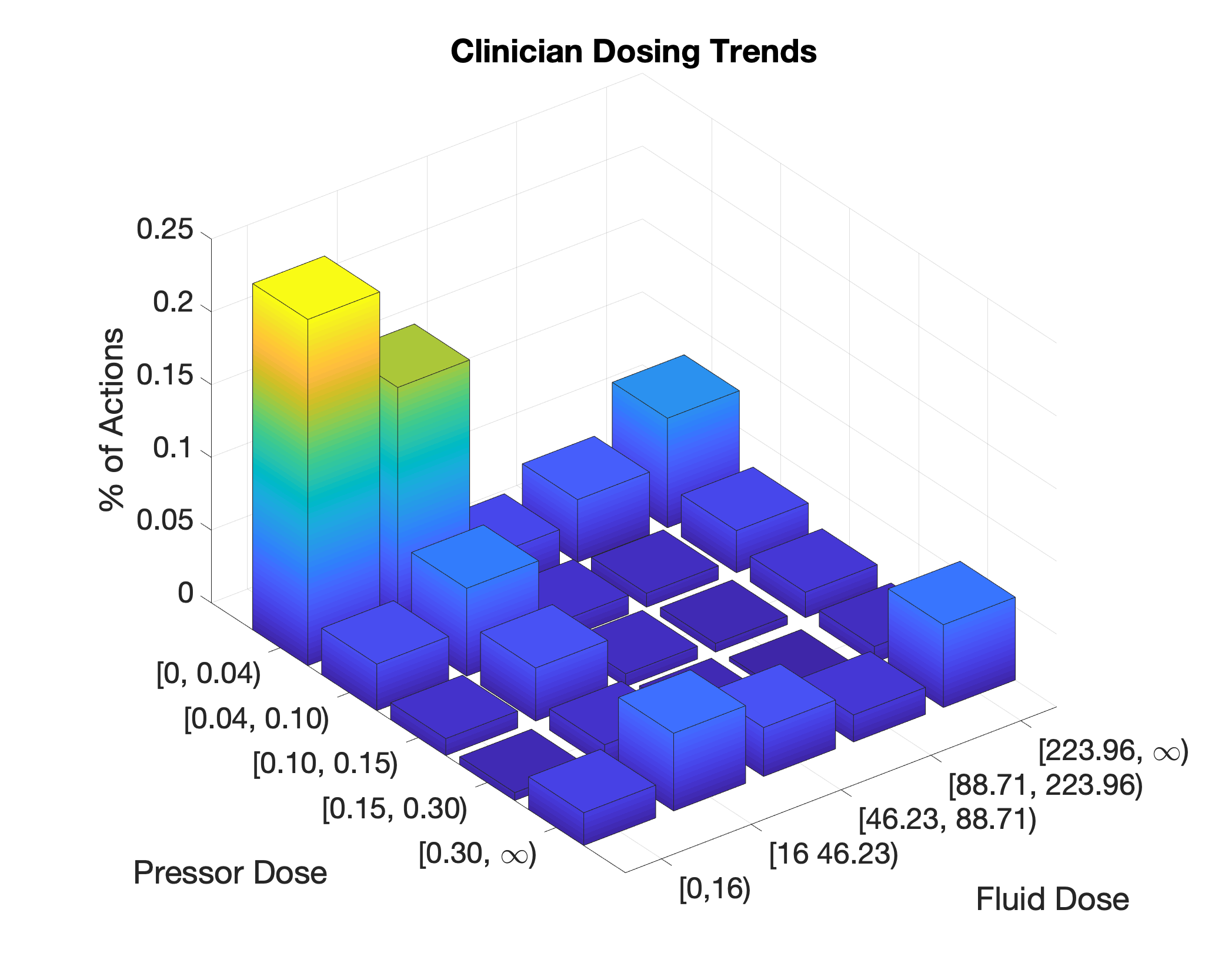}
\end{center}
\caption{Summary of the real clinician actions in our MIMIC III cohort.}
\label{Figure: Clinician Action Summary}
\end{figure}

\begin{figure}[h]
\begin{center}
\includegraphics[width = 0.75\textwidth]{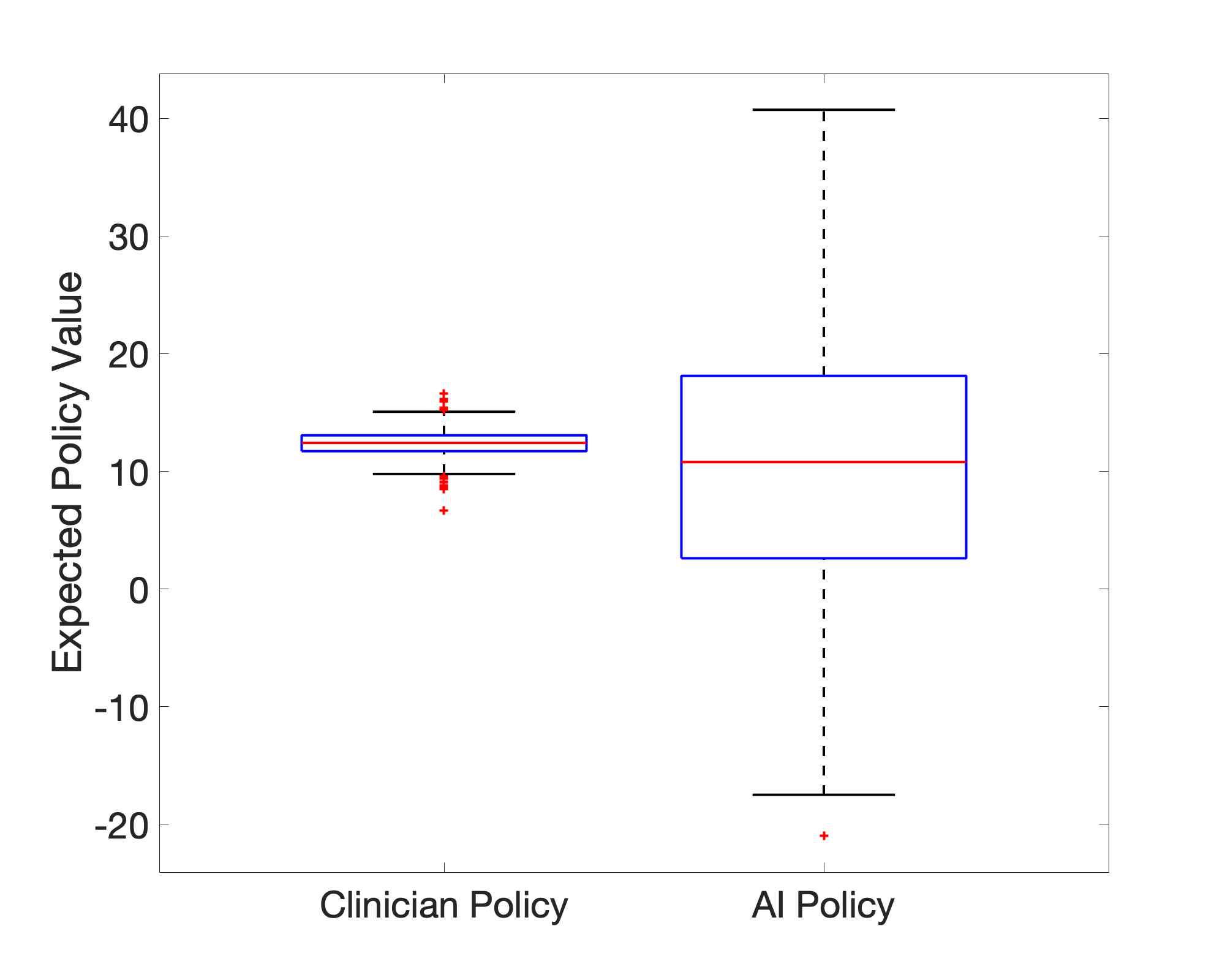}
\end{center}
\caption{Comparison of the expected policy value for the clinician policy and AI policy for 500 realizations of the environment. The lower expected policy values arise because our cohort is inherently more sick (our cohort has a mortality rate of 40\%). Given the reward for 90-day survival is +100 and the reward for death is -100, 40\% mortality in the population roughly corresponds to an expected policy value of 15 - 20. The best AI policy gives an expected policy value of ~40, which roughly corresponds to a 30\% mortality ((70\% x 100 ) - (30\% x -100)).}
\label{Figure: Our Model Value Comparison}
\end{figure}

In terms of expected value, the 95-percentile of AI Clinician policy performs much better than the actual clinicians, but what does this AI Clinician's policy look like? Below we show the AI Clinician (labeled ``RL Agent'') and an actual clinician's dosing suggestions for a given patient throughout their clinical course plotted alongside the patient's Mean Arterial Pressure (MAP).

\begin{figure}[h]
\begin{center}
\includegraphics[width = 0.75\textwidth]{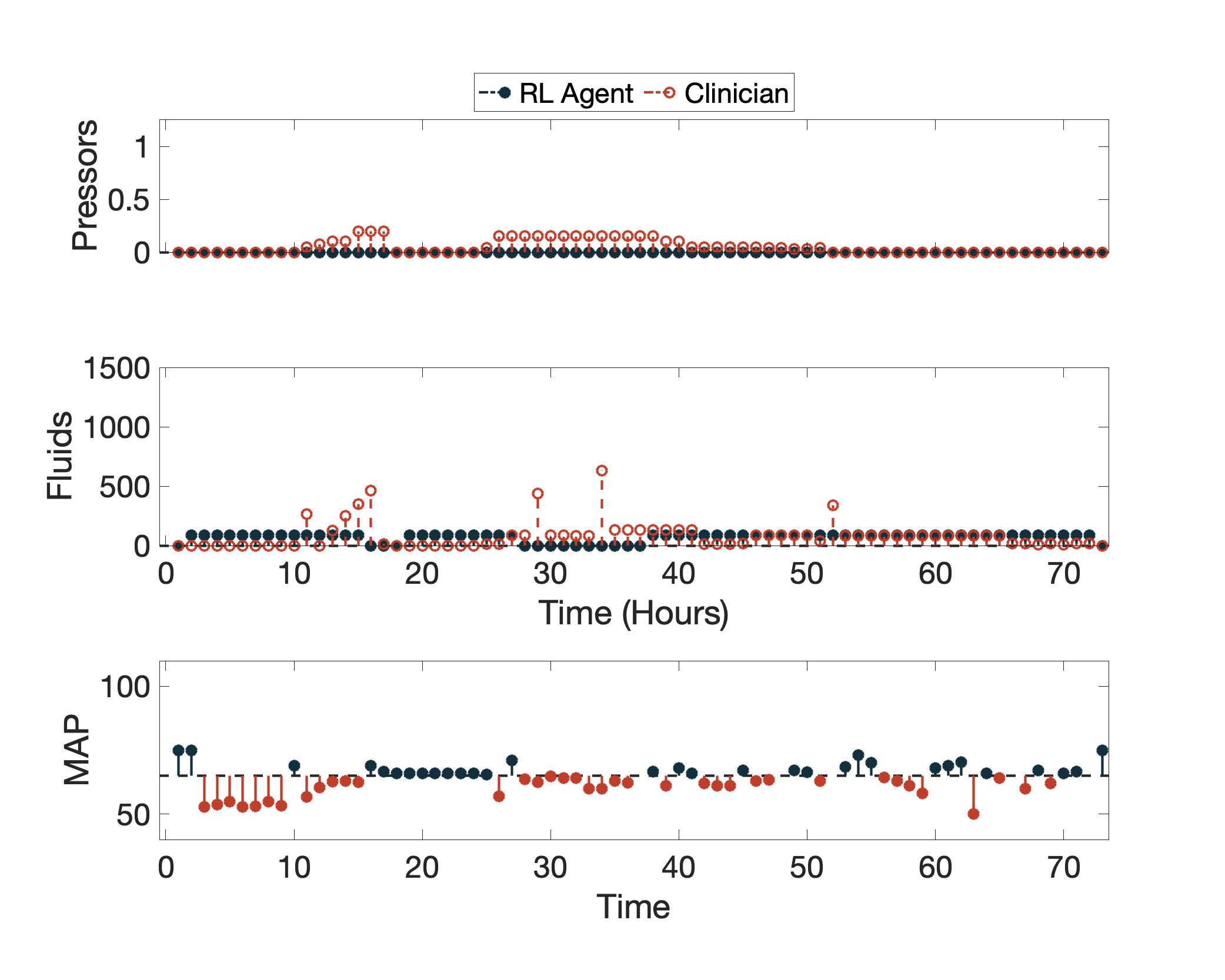}
\end{center}
\caption{Comparison between the AI Clinician's recommendations at a given point and the clinician's actual medication doses. The RL agent (AI Clinician) seems to be reluctant to to dose this patient (see hours 20-35), but why?}
\label{Figure: Our Real Action Recommendations}
\end{figure}

Despite having high expected return values, the AI Clinician fails to learn the relationship between MAP and the need for vasopressors and IV fluids (see hours 20 through ~35). Notice that for a significant portion of time the patient's MAP is below the recommended target of 65 mm Hg \cite{asfar_high_2014}, and the AI Clinician does not recommend any treatment. 
So, why does the AI Clinician yield such high expected return, yet misses several hours of critical hypotensive episodes? In our recreation of their work, we believe to have found the answer: \textit{Both the representation of the system (patient state-space and clinician actions) and how the policies are evaluated (the particular method of counterfactual reasoning) skew the expected returns of the AI Clinician.}

\section{Policy Evaluation}
\subsection{Importance Sampling Introduces Bias in Evaluating the AI policy in Patients with Poor Outcomes}

Komorowski et al. measure the expected value of the AI Clinician’s policy using importance sampling. In very simple terms this means that certain input parameters in a simulation have more significance than others . In the context of the AI Clinician's training, actions that closely matched a clinician's and resulted in good outcomes were given the highest weight in training. This seems like a reasonable approach; however, it has the effect of discounting patterns learned from medically complex patients who deteriorated despite the clinical teams best attempt to save the patient (remember 90-day survival is a very challenging goal to meet for some very sick ICU patients). This approach also has the perverse effect of disproportionately rewarding simple decisions, like taking no action on a stable patient.

The AI Clinician is built from retrospective data and was never able to directly interact with a real patient environment to determine its effectiveness. Traditionally, Reinforcement Learning works by allowing the agent (in our case, the AI Clinician) to act on the environment and learn optimal strategies via trial-and-error, an approach that would be patently unethical if applied prospectively to real patients. To train the AI Agent Komorowski et al. created a ``simulated ICU'' (i.e. the \emph{in silico} environment, see Summary). The AI Clinician is then graded on its performance in this oversimplified \emph{in silico} ICU environment. 

Importance sampling, as used by Komorowski et al., represents a middle ground between the above concepts, the grading the AI Clinician performance in their created \emph{in silico} ICU and its prospective performance in a real ICU. Trajectories are navigated using the real clinician’s policy, and the most weight is given to the cases in which the AI Clinician’s decisions most closely approximated the real clinician’s decisions. In the final policy evaluation, almost no value (positive or negative) is given to trajectories for which the AI Clinician tries something significantly different than the real clinician, regardless of if it would have harmed or improved a patient's outcome. Unfortunately, some patients have poor outcomes and die despite optimal medical management by human clinicians, and these events are taken into account when evaluating the clinician's policy and are ignored when evaluating the AI policy; since the AI agent can learn to avoid dosing patients who are very sick, or the AI agent can learn to ``game the system'' by just acting differently than the clinician in tough cases.

As a consequence of importance sampling, and due to a lack of understanding of causal links, the AI Clinician learns the association between the aforementioned interventions and the poor outcome, and thus avoids recommending these optimal clinical actions in order to not be penalized for the death of the patient. This introduces significant bias when estimating the expected value of the AI Clinician’s policy as the AI Clinician does not get graded on the tough cases. 

\subsection{Limitations of Long-Term Rewards}

IV fluids and vasopressors are used to treat hypotensive episodes in septic patients over a short period of time, but these interventions do not treat the underlying causes of sepsis. In general, clinical decision-making strikes balance between short-term (e.g. maintaining a physiologic MAP or appropriate hourly urine output) and long-term outcomes (e.g. preventing kidney injury or mortality). The approach used by Komorowski et al. focuses \emph{only} on long-term outcomes: 90-day mortality, or in-hospital mortality if they do not have the 90-day mortality data. The work \emph{entirely ignores} short-term rewards (e.g. maintaining physiologic blood pressure) preventing the AI Clinician from learning the crucial relationship between IV fluid and vasopressor dosing and MAP. Omitting intermediate rewards ultimately allows the AI Clinician to learn non-standard clincical behaviors that may have deleterious consequences (e.g. giving vasopressors to a normotensive patient). Ironically these authors raise their disdain for omitting intermediate rewards in RL work in a recent commentary that was also published in \emph{Nature Medicine} \cite{norgeot_call_2019}.

\section{System Representation}
\subsection{Trajectory Discretization}
The authors reduce all of the patient information into four hour summaries or ``bins.'' This is adequate for static variables (like race and co-morbid conditions), and even some dynamic laboratory variables like leukocyte count, which does not change dramatically minute-to-minute, or even hour-to-hour. However, in septic patients some variables like heart rate, blood pressure, and urine output can undergo rather impressive changes over the course of four hours and patients undergoing rapid deterioration in their health are the ones in most need of treatment. Down-sampling patient data into four hour bins severely limits the AI Clinician's ability to detect and respond to these episodes of deterioration.

Vasopressor therapy is a highly reactive intervention. Vasopressors work by constricting the blood vessels and increasing the contractility of the heart to increase blood pressure. If a clinician has a septic patient with low MAP (e.g. below 65 mm Hg \cite{asfar_high_2014}), they will consider putting them on vasopressors to increase blood pressure. The dose-to-response time of vasopressors is nearly instantaneous. The four-hour bins, utilized by Komorowski et al., fail to adequately capture how quickly this process happens, and does not capture the relationship between states and actions. 

Within a four-hour bin, it is possible for a patient’s MAP to plummet, a clinician to intervene with vasopressors, and MAP to recover. The summary for this bin could show that the patient has normal MAP, yet was administered vasopressors. The AI Clinician could learn from this state and action that it is optimal to give vasopressors to patients with normal MAPs.

\subsection{States, Actions, and Transitions}

As mentioned above, the authors map four-hours of patient observations to one of the 750 pre-defined states. That is, each combination of heart rate, blood pressure, and pre-existing conditions are reduced to 750 states. Transition matrices created in this way are incapable of capturing the temporal trends required for assessing the treatment needs of a patient, as they only provide information about the current and next state. The implication of this design is that low frequency events (i.e. rapid decompensation) are all but ignored by the AI Clinician.

The proposed \emph{in silico} model is biased towards healthier patients which comprise nearly two thirds (Figure 8) of the populations in both cohorts. Together with the importance sampling-based evaluation metrics, this explains why the zero drug policy appears to be such a successful dosing policy.

\section{Interpretability}

The authors state that their proposed model is interpretable, however, a close examination reveals that the ``interpretability'' analysis presented by the authors is misleading at best. Supplementary Figure 2 caption describes the interpretability method used by Komorowski et al: 
\begin{quote}
    We built classification random forest models to predict whether the medications were (clinicians’ policy) or should have been (AI policy) administered (regardless of the dose), using patient variables as input data $\ldots$ . Then, the relative importance of each variable was estimated $\ldots$ .This confirmed that the decisions suggested by the reinforcement learning algorithm were clinically interpretable and relied primarily on sensible clinical and biological parameters $\ldots$ 
\end{quote}
This analysis is flawed on at least three different levels:

\subsection{The Proposed Method has Nothing to do with Interpretability}
The proposed method only reveals the top factors contributing to the discrepancy between the clinical policy and the AI clinician policy. It says nothing about the interpretability of the individual dosing decisions made by the AI clinician. To elucidate why this distinction is important, let us consider a hypothetical scenario. Consider a new policy (policy-X) that follows the clinical policy when the patient has a low to moderate comorbidity index \cite{charlson_validation_1994}, and withdraws treatments when the patient has a high comorbidity index. In this case, the interpretability model proposed by the authors simply reveals the top factors contributing to the comorbidity index, and on the subset of patients with low to moderate comorbidity index the method reveals nothing about the clinical rationale for the treatments proposed by the clinical policy and/or the policy-X.

\subsection{Random Forest Gives Random Variations in Importance}
The authors provide separate ``feature importance'' plots for the clinician's policy and AI policy. In a binary classification task, these two plots should be essentially identical. The differences that we observe has little to do with the ``feature importance,'' and is purely an artifact of utilizing a random forest model. 

\subsection{Individual Actions Are Not Interpretable}
Assuming all the analysis was performed correctly (and assuming no other confounding factor factors affecting the dosing policy), still the method of ``feature importance'' says nothing about the AI agent's reasoning for the individual dosing recommendations. At best, this method is a global measure of interpretability (over all time points and all patients).

\section{Externally Validated?}

The importance of externally validating a machine learning model on a data set different than the training cohort is an absolute minimum if one hopes to claim that the work can be generalized to other institutions. It shows that the model is robust and does not over fit a single data set. 

Komorowski et al. address the issue of external validation by considering a cohort of 3.3 million ICU patients provided by their industry partner, Philips Healthcare, a medical device manufacturer with a division dedicated to monetizing healthcare analytics in Baltimore, Maryland. 

Close inspection of the data sets used in this work creates almost as many questions as it answers. We have listed our top concerns below in no particular order; however, we leave the reader to arrive at their own conclusion on whether the eRI dataset provided by Philips Healthcare really provided a valid validation dataset for this study.

\subsection{Sepsis Incidence}
What is the incidence of sepsis in the 3.3 million Philips Healthcare patients? It is impossible to determine this from the data seen provided by the authors. Using the numbers they provide in Figure 6 one gets the absurd value of 83.5\% for the incidence of sepsis, which is clearly due to their biased exclusion criteria. However, from the data presented, it is not possible to calculate the incidence of sepsis in the cohort. For the MIMIC-III data set one can determine that it had an incidence of 31.4\% which is also higher than expected, but much more reasonable. Furthermore, the authors admit to using a ``modified definition'' of sepsis to determine inclusion in the final cohort. These two issues make it very challenging to evaluate the validity of the eRI cohort.

\begin{figure}[h]
\begin{center}
\includegraphics[width = \textwidth]{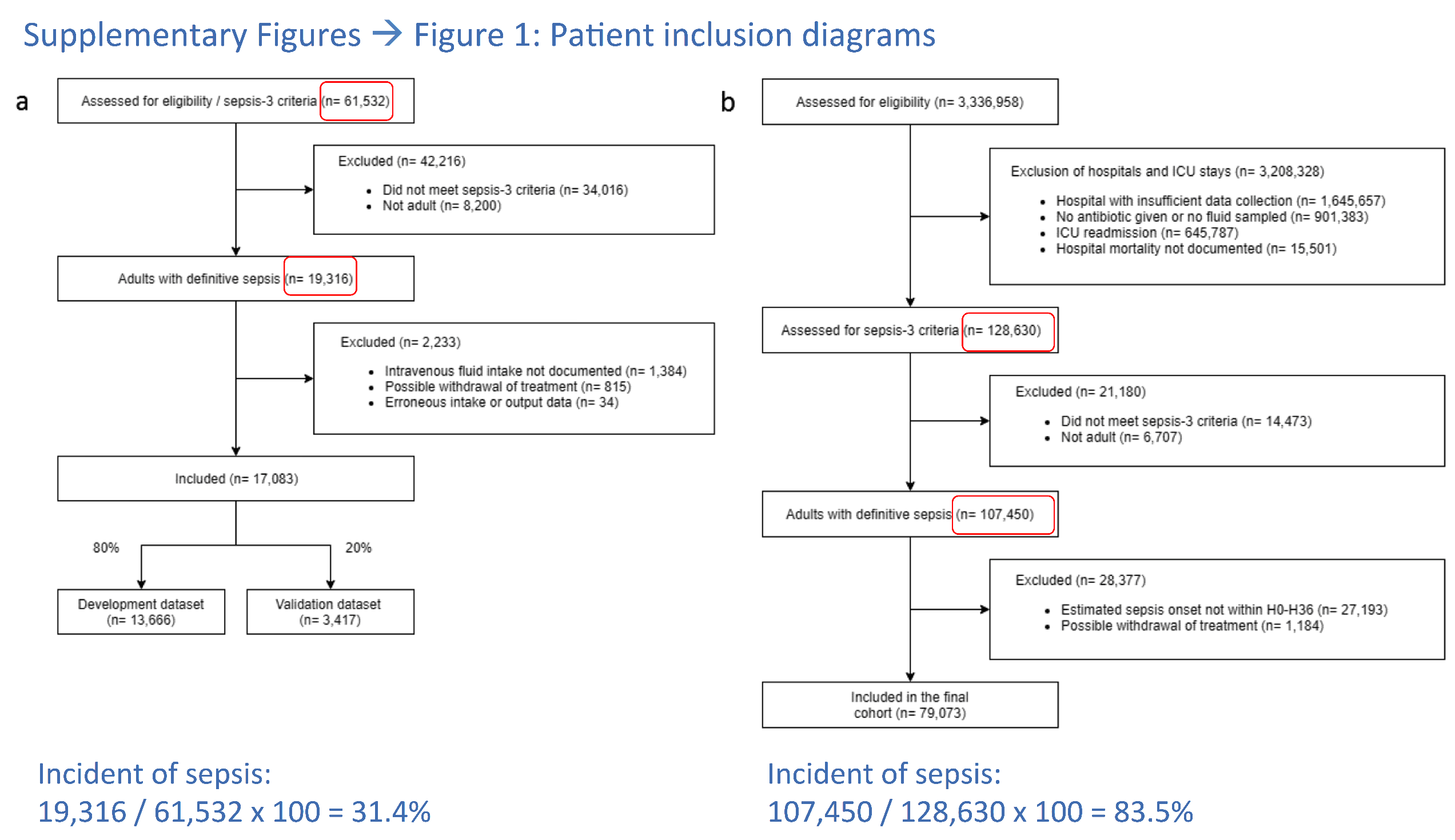}
\end{center}
\caption{Comparison of the MIMIC cohort (left) and the eRI cohort (right). Figure taken from appendix of \cite{komorowski_artificial_2018}.}
\label{Figure: Cohort Comparison}
\end{figure}

\subsection{Data Quality}

In describing the Philips Healthcare data set the authors state that “given that our analysis resolution was 4 h, we expected at least six records [of fluids/vasopressors] per day.” Since these records are counted before binning, this begs the question whether these six records could have been taken all within a given time bin (or within an hour for that matter), leaving the remaining 5 time bins (20 hours) with no records of vasopressors and fluids. In contrast, the quality of data in the MIMIC database does seem to be higher with ``20.4 intravenous fluids records and 31.1 vasopressor records [per day].'' 
When one recognizes that 1.66 million patients (49\%) from the original data set were excluded from consideration in round one (Figure 6) for missing data it seriously calls into question the quality of the remaining data (only 2.3\% of the data made it into the final analysis).

\subsection{AI Fails in the Philips Healthcare Dataset}

When the AI Clinician operates inside the MIMC-III data set there appears to be expected punishments for both under-dosing and over-dosing fluids (Figure 7, Top Left). However when the AI Clinician begins to treat patients in the Philips Healthcare dataset (Figure 7, Bottom Left) it appears as though under-dosing patients with IV fluids by as much as 1.2 L, has no effect on the mortality, while it nearly doubles the mortality rate within the MIMIC database. The substantial difference between the dose response to vasopressors between the two cohorts is also concerning, and leads one to question whether or not the AI Clinician actually has similar performance between the two datasets.

It is worth noting that using these curves alone (Figure 7), one cannot evaluate the AI Clinician's performance, as an individual dose-mortality pair does not take into account confounding factors, such as baseline comorbidities and whether or not a patient received antibiotics in a timely manner. For example, a patient who gets 800 mL of extra fluid (over a 4 hour window) compared to the AI Clinician may have been sicker at the baseline, and the AI Clinician may have decided to withdraw treatment (see the section Evaluation Bias).

\begin{figure}[h]
\begin{center}
\includegraphics[width = 0.9\textwidth]{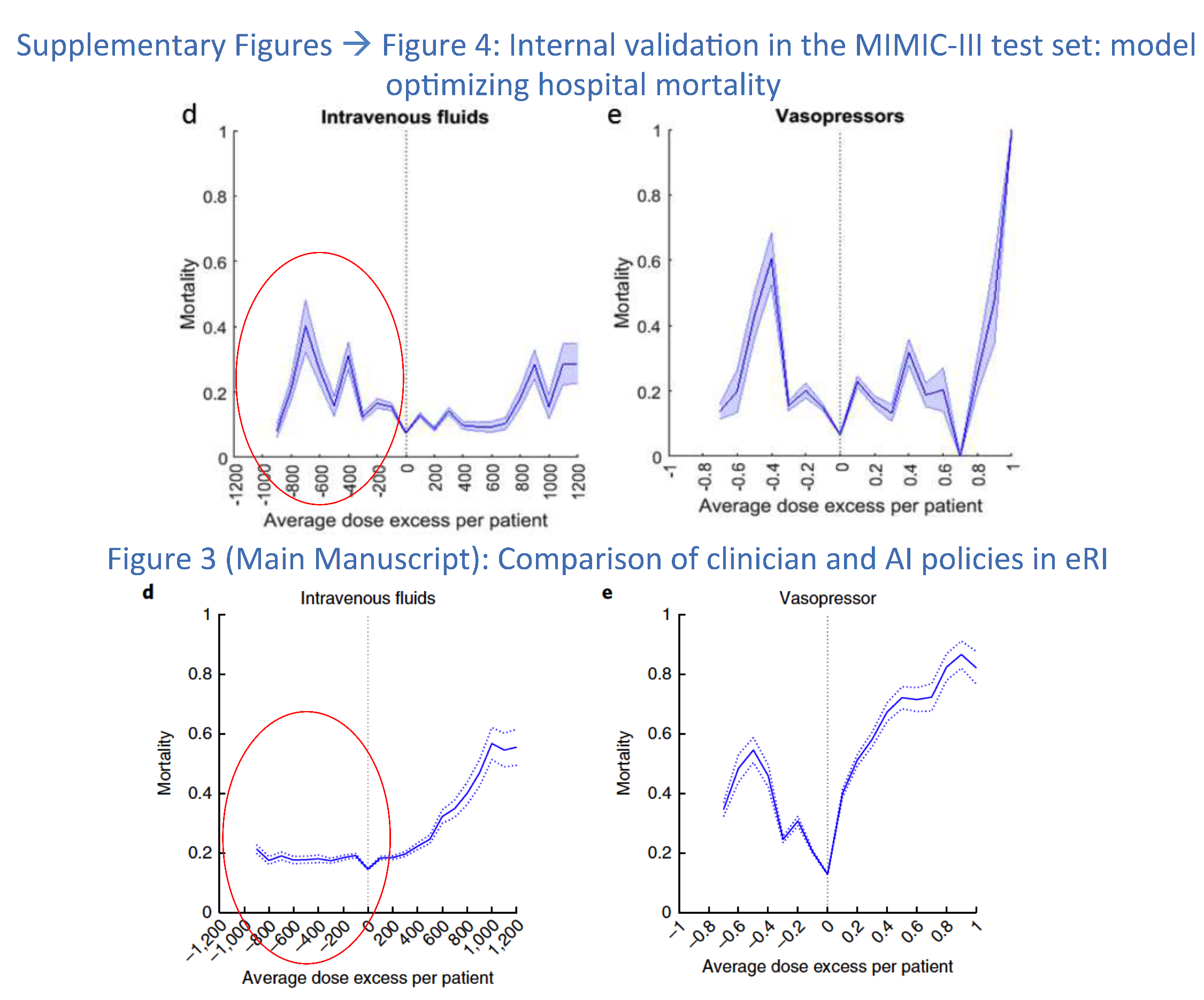}
\end{center}
\caption{Comparison of the MIMIC (top) and eRI (bottom). }
\label{Figure: AI Cohort Comparison}
\end{figure}

\section{The Wrong Cohort Entirely}

The authors train and evaluate the AI Clinician on patients that meet Sepsis-3 criteria from two different data sets, but is this the right group to evaluate? At point85 we'd say no for the two reasons listed below.

\subsection{Unnecessary AI}

As we discuss previously, the method of evaluating the AI Clinicians performance disproportionately rewards actions that are different from the clinician (often inaction when the clinician is significantly intervening). As seen in Figure 8 only one-third of either cohort ever received a vasoactive drug, and from this information one could safely assume that only one-third of the patient cohort experienced clinically significant hypotension. The take home point is that of all the patients that Komorowski et al. examined, nearly two-thirds received no vasopressor treatments in their ICU course, meaning that two-thirds of the patients had no need for the AI Clinician's intervention. It is not surprising that the ``zero drug'' policy then had such a high expected reward.

\subsection{Iatrogenic Treatment}

To date, all evidence-based practices regarding the treatment of sepsis reserve the use of vasopressors to those patients suffering from hypotension, invariably defined as a MAP $<$ 65. It would be considered universally iatrogenic to administer a normotensive patient vassopressors. Unfortunately, there is no guarantee that the AI Clinician won't make this mistake. The lack of intermediate rewards and the use of four-hour binning previously described means that \emph{there is ample opportunity for the AI Clinician to have learned to give normotensive patients vasopressors.}

\begin{figure}[h]
\begin{center}
\includegraphics[width = \textwidth]{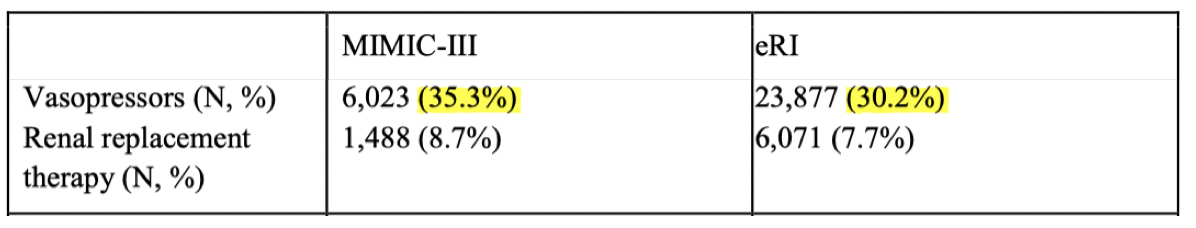}
\end{center}
\caption{One can safely assume that only about one-third of either cohort warranted any consideration by the AI Clinician, meaning that for two-thirds of the either cohort the AI Clinician prowess was rewarded as it recommended doing nothing.}
\label{Figure: Cohort Vasopressor Needs}
\end{figure}

\section{Final Comments}

We have gone to great lengths to fairly examine and reproduce this work. The corresponding authors have yet to provide access to their source code, which we requested in October 2018. Studies such as the work by Komorowski et al. and academic journals seeking to retain a dwindling readership through the hasty promotion of impressive but unvetted claims are contributing to an erosion of public trust and are precipitating the ``big-data winter'' cautioned by Michael Jordan.

``Before the Second World War meteorology had been a bit like medicine in the nineteenth century: 
\emph{the demand for expertise was so relentless that the supply had no choice but to make fraudulent appearances.}''
-\textit{The Coming Storm}\cite{lewis2018coming}

On 2/4/19 Dr. Faisal Aldo provided a link to the GitHub repository containing MatLab code necessary for reproducing parts of their work. We will attempt to apply this code to the cohort that we selected for this post; however, reproduction of the exact cohort used by Komorowski et al. without more specific guidance may not be possible. The most time consuming step in this field is pre-processing a dataset for evaluation by the algorithm (e.g. how were missing values handled, which of the many, many infusions in MIMIC-III were selected \& which were discarded?).

The code generated by Jeter et al. that accompanies this paper can be found here:

\url{https://github.com/point85AI/Policy-Iteration-AI-Clinician}
\bibliographystyle{unsrt}
\bibliography{Elmo}

\begin{thebibliography}{10}

\bibitem{komorowski_artificial_2018}
Matthieu Komorowski, Leo~A. Celi, Omar Badawi, Anthony~C. Gordon, and A.~Aldo
  Faisal.
\newblock The {Artificial} {Intelligence} {Clinician} learns optimal treatment
  strategies for sepsis in intensive care.
\newblock {\em Nature Medicine}, October 2018.

\bibitem{timmer_ais_2018}
John Timmer.
\newblock {AIs} trained to help with sepsis treatment, fracture diagnosis,
  October 2018.

\bibitem{sutton_reinforcement_nodate}
Richard~S Sutton and Andrew~G Barto.
\newblock Reinforcement {Learning}: {An} {Introduction}.
\newblock page 352.

\bibitem{alzantot_deep_2017}
Moustafa Alzantot.
\newblock Deep {Reinforcement} {Learning} {Demysitifed} ({Episode} 2) —
  {Policy} {Iteration}, {Value} {Iteration} and…, July 2017.

\bibitem{johnson_mimic-iii_2016}
Alistair~E.W. Johnson, Tom~J. Pollard, Lu~Shen, Li-wei~H. Lehman, Mengling
  Feng, Mohammad Ghassemi, Benjamin Moody, Peter Szolovits, Leo Anthony~Celi,
  and Roger~G. Mark.
\newblock {MIMIC}-{III}, a freely accessible critical care database.
\newblock {\em Scientific Data}, 3:160035, May 2016.

\bibitem{besag_exact_2013}
J.~Besag and D.~Mondal.
\newblock Exact {Goodness}-of-{Fit} {Tests} for {Markov} {Chains}: {Exact}
  {Goodness}-of-{Fit} {Tests} for {Markov} {Chains}.
\newblock {\em Biometrics}, 69(2):488--496, June 2013.

\bibitem{asfar_high_2014}
Pierre Asfar, Ferhat Meziani, Jean-François Hamel, Fabien Grelon, Bruno
  Megarbane, Nadia Anguel, Jean-Paul Mira, Pierre-François Dequin, Soizic
  Gergaud, Nicolas Weiss, François Legay, Yves Le~Tulzo, Marie Conrad, René
  Robert, Frédéric Gonzalez, Christophe Guitton, Fabienne Tamion, Jean-Marie
  Tonnelier, Pierre Guezennec, Thierry Van Der~Linden, Antoine Vieillard-Baron,
  Eric Mariotte, Gaël Pradel, Olivier Lesieur, Jean-Damien Ricard, Fabien
  Hervé, Damien du~Cheyron, Claude Guerin, Alain Mercat, Jean-Louis Teboul,
  and Peter Radermacher.
\newblock High versus {Low} {Blood}-{Pressure} {Target} in {Patients} with
  {Septic} {Shock}.
\newblock {\em New England Journal of Medicine}, 370(17):1583--1593, April
  2014.

\bibitem{norgeot_call_2019}
Beau Norgeot, Benjamin~S. Glicksberg, and Atul~J. Butte.
\newblock A call for deep-learning healthcare.
\newblock {\em Nature Medicine}, 25(1):14--15, January 2019.

\bibitem{charlson_validation_1994}
Mary Charlson, Ted~P. Szatrowski, Janey Peterson, and Jeffrey Gold.
\newblock Validation of a combined comorbidity index.
\newblock {\em Journal of Clinical Epidemiology}, 47(11):1245--1251, November
  1994.

\bibitem{lewis2018coming}
Michael Lewis.
\newblock {\em The Coming Storm}.
\newblock Audible Studios, 2018.

\end{thebibliography}

\end{document}